\documentclass{ifacconf}

\usepackage{natbib}        
\usepackage{balance}  
\usepackage{algorithm}
\usepackage{bm}
\usepackage{amsmath}
\usepackage{amsfonts}
\usepackage{mathrsfs}
\usepackage{algorithm}
\usepackage{algpseudocode}
\usepackage{graphicx}
\usepackage{framed}
\usepackage{subcaption}
\usepackage{array}
\usepackage{booktabs}
\usepackage{wrapfig}
\usepackage{eucal}
\usepackage{flushend}
\usepackage{comment}

\begin{document}
\begin{frontmatter}

\title{Hybrid Task Constrained Planner for Robot Manipulator in Confined Environment} 


\author[First]{Yifan Sun} 
\author[First]{Weiye Zhao} 
\author[First]{Changliu Liu}

\address[First]{Robotics Institute, 
   Carnegie Mellon University, Pittsburgh, PA 15213 USA (e-mails: yifansu2, weiyezha, cliu6@andrew.cmu.edu).}

\begin{abstract}               
  Trajectory generation in confined environment is crucial for wide adoption of intelligent robot manipulators. In this paper, we propose a novel motion planning approach for redundant robot arms that uses a hybrid optimization framework to search for optimal trajectories in both the configuration space and null space, generating high-quality trajectories that satisfy task constraints and collision avoidance constraints, while also avoiding local optima for incremental planners. Our approach is evaluated in an onsite polishing scenario with various robot and workpiece configurations, demonstrating significant improvements in trajectory quality compared to existing methods. The proposed approach has the potential for broad applications in industrial tasks involving redundant robot arms.
\end{abstract}


\end{frontmatter}

\section{Introduction}
\vspace{-10pt}
In recent years, intelligent robot manipulators have become increasingly common in industrial and daily life settings. However, generating high-quality trajectories remains a significant challenge due to the high dimensionality of the configuration space and various constraints, such as kinematic, dynamic, safety, and task constraints (\cite{he2023hierarchical}). Safety constraints require the robot to maintain a safe distance from humans and workpieces to avoid collisions and injuries, while task constraints vary depending on the specific scenario. In confined industrial spaces, it is difficult to solve the feasible inverse kinematics (IK) problem that satisfies both safety and task constraints in the configuration space, which is especially hard for robot with high degree of freedom (DOF).

An efficient approach to address the complex trajectory generation problems is to break them down into step-wise subproblems, known as incremental planning (\cite{zhao2020contact}). However, this approach heavily relies on the quality of the reference configuration from the previous time step, making it vulnerable to getting stuck in local optima with bad initializations. On the other hand, by exploiting the additional degrees of freedom in the null space, redundant robots can achieve the same position and orientation with various joint configurations, which could help avoid local optima. 

In this paper, we present a hybrid task constrained planner framework that generates high-quality trajectories efficiently while satisfying safety and task constraints. The framework uses an incremental planning scheme, convexifies non-convex inequality constraints, and linearizes nonlinear equality constraints. To overcome local optima, the framework searches for configurations in the null space with respect to task constraints while maximizing a safety-oriented objective. Our framework can generate desired trajectories that satisfy all constraints in real-time and overcome local optima challenges for incremental planners. The contributions of this paper are summarized as follows:
\begin{enumerate}
    \item We introduce the Hybrid Task Constrained Planner (HTCP) framework, which utilizes iterative convex optimization and null space guided search to generate optimal task-constrained trajectories efficiently.
    \item Our method is tested extensively with various task configurations in real-world robot polishing experiments, demonstrating its feasibility and robustness in industrial scenarios.
\end{enumerate}

\section{Related Work}

There are three main categories of motion planning methods: sample-based methods, energy function-based methods, and optimization-based methods.

Sample-based methods, such as probability road maps (PRMs) (\cite{kavraki1996probabilistic}) and rapidly-exploring random trees (RRT) (\cite{lavalle1998rapidly}), randomly sample configuration space to generate collision-free trajectories. The Constrained Bi-directional Rapidly-exploring Random Tree (CBIRRT) (\cite{berenson2009manipulation}) can handle task constraints by planning paths in the manifold of configuration space. However, sample-based methods may become inefficient for higher-dimensional configuration spaces, resulting in trajectories that are not smooth or accurately satisfying the task equality constraints.


Energy function-based methods (\cite{zhao2021model, zhao2022probabilistic}), such as potential field methods (\cite{khatib1986real}), control barrier functions (\cite{Chiara2019ECC}), and safe sets (\cite{zhao2022safety}), use an energy function to attract the robot to the target position and repel it from obstacles. These methods can be extended to tasks with redundant DOFs by applying the potential field to generate the velocity in the null space movement. However, these methods are prone to getting stuck at local optima and may not accurately satisfy task constraints due to the highly nonlinear nature of the energy function.

Optimization-based methods generate trajectories by formulating an optimization problem with an objective function and multiple constraints. CHOMP (\cite{ratliff2009chomp}) and ITOMP (\cite{park2012itomp}) algorithms are examples that solve motion planning problems with different penalty functions for joint velocities, accelerations, and distance from obstacles. These methods generate smoother trajectories and handle both safety inequality and task equality constraints. However, the highly nonlinear and non-convex nature of the optimization problem makes it computationally expensive to obtain a solution using generic nonlinear optimization solvers. To address this challenge, various methods directly convexify the optimization problem using domain knowledge to efficiently handle non-convex inequality constraints and nonlinear equality constraints. The Convex Feasible Set (CFS) algorithm iteratively convexifies constraints (\cite{liu2017convex}), the iterative LQR algorithm linearizes constraints (\cite{Emanuel2004ICINCO}), and Iterative Convex Optimization for Planning (ICOP) (\cite{zhao2020contact}) absorbs both advantages from CFS and iLQR to generate high-quality trajectories incrementally in real-time. Optimization-based methods can provide good global solutions with acceptable real-time performance, but rely on the quality of the initialization and are subject to the problem of local optima.

\section{Problem Formulation}
This paper  focuses on contact-rich trajectory generation in confined environment. The robot state is denoted as $x \in X \subset \mathbb{R}^N$, where $X$ is the configuration (state) space and  $N$ is its dimension, which specifies the degree of freedom of the robot. The robot state at a discrete time step $t$ is denoted as $x_t$. A trajectory is defined to be the sequence of states from time $1$ to time $T$: $\bold{x} = [x_1;x_2;...;x_T] \in \mathbb R^{NT}$.

\subsection{Task Constraint}
The contact-rich task equality constraints are defined so that the robot body is in contact with the specified targets. Mathematically, the constraint can be written as:
\begin{equation}
\Gamma_j(x_t) = p_j^t,
\label{2}
\end{equation}
where the $\Gamma_j(\bullet): \mathbb{R}^N \to \mathbb{R}^M $ is a generalized function to project the $j$-th point on the robot body (e.g., tool tip) to the constrained task space given the robot configuration $x_t$. $M\in\{1,2,\ldots, 6\}$ is the dimension of the task space. $M=2$ means the task is constrained in a 2-dimensional plane, e.g., wiping a surface. $M=3$ means the task constraints are defined in a 3-dimensional Cartesian space, e.g., welding on a predefined trajectory. This paper mainly considers the case $M=3$ to construct the task constraints with redundant DOFs for the robot.

\subsection{Safety Constraint}

Suppose that the workspace is occupied by obstacles in the Cartesian space, denoted by $O\subset\mathbb{R}^3$. The safety constraint for the robot at time step $t$, with state $x_t$, is defined as:
\begin{equation}\label{eq: safety}
D(x_t, O) > 0,
\end{equation}
where $D(\bullet): \mathbb{R}^N \times \mathbb{R}^3 \to \mathbb{R}$ is a signed distance function that computes the distance from the robot to the obstacle in Cartesian space. In order to ensure safety, it is necessary to ensure that the distance is greater than zero.

\subsection{Optimization Objective}

The objective of planning in a confined environment is to generate smooth trajectories that satisfy both safety inequality constraints and task equality constraints. Mathematically, this objective can be formulated as the following optimization problem:
\begin{equation}
\begin{aligned}
& \underset{\bold{x}}{\text{min}}
& & \bold{J}(\bold{x}) = \| \bold{x}\|_Q^2 \\
& \text{s.t.} & &  \bold{\Gamma}_j(\bold{x}) = \bold{p}_j, \forall j = 1,2,3,...,\\
& & &  \bold{D}(\bold{x}, O) > 0,\\
& & &  x_{min} \leq x_t \leq x_{max},\forall t = 1,2,3,...,T\\\\
\end{aligned}
\label{eq:opt}
\end{equation}
In this formulation, $x_{min}$ and $x_{max}$ are the joint limits, and $\| \bold{x}\|_Q^2 = \bold{x}^\top Q \bold{x}$, where $Q = V^\top V$ and $V \in \mathbb R^{N(T-1)\times NT}$ is a finite difference operator that extracts velocity of the trajectory. The objective function $\bold{J}(\bold{x})$ minimizes the sum of the norm difference between joint positions in consecutive time steps, allowing the robot to move smoothly. The constraints require that the robot maintains contact ($\bold{\Gamma}_j(\bold{x}) = \bold{p}_j$), ensures safety ($\bold{D}(\bold{x}, O) > 0$), and stays within joint limits ($x_{min} \leq x_t \leq x_{max}$). 



\subsection{Challenge}

\textbf{Challenge 1:}
Directly solving \eqref{eq:opt} is not a trivial task, as highly nonlinear and non-convex equality and inequality constraints must be satisfied for every time step. Instead of optimizing the trajectory as a whole, we choose to solve the state $x$ from \eqref{eq:opt} incrementally. This approach has been demonstrated as efficient and optimal in (\cite{zhao2020contact}). Mathematically, the optimization problem for solving $x_t$ can be formulated as:
\begin{subequations}
\begin{align}
& \underset{x_t}{\text{min}}
& & J_1(x_t,x_{t-1}) = \| x_t - x_{t-1}\|_Q^2 \\
& \text{s.t.} & &  \Gamma_j(x_t) = p_j^t, \forall j = 1,2,3,..., \label{eq: eqcon}\\
& & &  D(x_t, O) > 0, \label{eq: ineqcon}\\
& & &  x_{min} \leq x_t \leq x_{max},
\end{align}
\label{eq:opt1}
\vspace{-10pt}
\end{subequations}

\textbf{Challenge 2:}
As noted in (\cite{zhao2020contact}), incremental planning has a significant drawback in that the solution for $x_t$ is heavily dependent on the reference configuration $x_{t-1}$. Furthermore, existing optimization methods for \eqref{eq:opt1} are susceptible to becoming trapped in local optima, which can result in constraints not being satisfied or the objective being excessively large. This can cause the entire solving process to terminate prematurely and fail to produce a high-quality trajectory. To address this issue, our method exploits the robot's redundant DOFs and resolves the local optima problem of \eqref{eq:opt1} by solving an auxiliary optimization problem in the null space when the process becomes stuck. Mathematically,
\begin{subequations}
\begin{align}
& \underset{x_t}{\text{min}}
& & J_2(x_t,x_{t-1}) = -D(x_t, O) + J_1(x_t,x_{t-1}) \\
& \text{s.t.} & &  \Gamma_j(x_t) = p_j^t, \forall j = 1,2,3,..., \label{eq: new eq}\\
& & &  x_{min} \leq x_t \leq x_{max},
\end{align}
\label{eq:opt2}
\vspace{-10pt}
\end{subequations}

The auxiliary optimization problem, unlike the original problem \eqref{eq:opt1}, converts the safety constraint into the objective function $J_2$. If problem\eqref{eq:opt1} stucks, indicating that a nearby configuration that satisfies all the constraints cannot be found, the auxiliary problem enables the robot to explore a new configuration while prioritizing obstacle avoidance to escape the local trap. By solving the original and auxiliary problems iteratively, the planner can generate a high-quality trajectory without sacrificing efficiency.

\section{Methodology}

In this section, we present the methodologies used to solve the optimization problem \eqref{eq:opt}. We begin by introducing the iterative configuration space optimization method, which aims to solve \eqref{eq:opt1} by linearizing the nonlinear equality constraints and convexifying the nonconvex inequality constraints. Next, we discuss the iterative null space optimization technique, which handles \eqref{eq:opt2} by exploring solutions from the null space defined by the equality constraint. Finally, we introduce the hybrid task constrained planner, which combines both optimization approaches to leverage the benefits of each, to efficiently generate high-quality solutions towards problem \eqref{eq:opt}. 

 \begin{figure}
 \centering
  \includegraphics[scale=0.35]{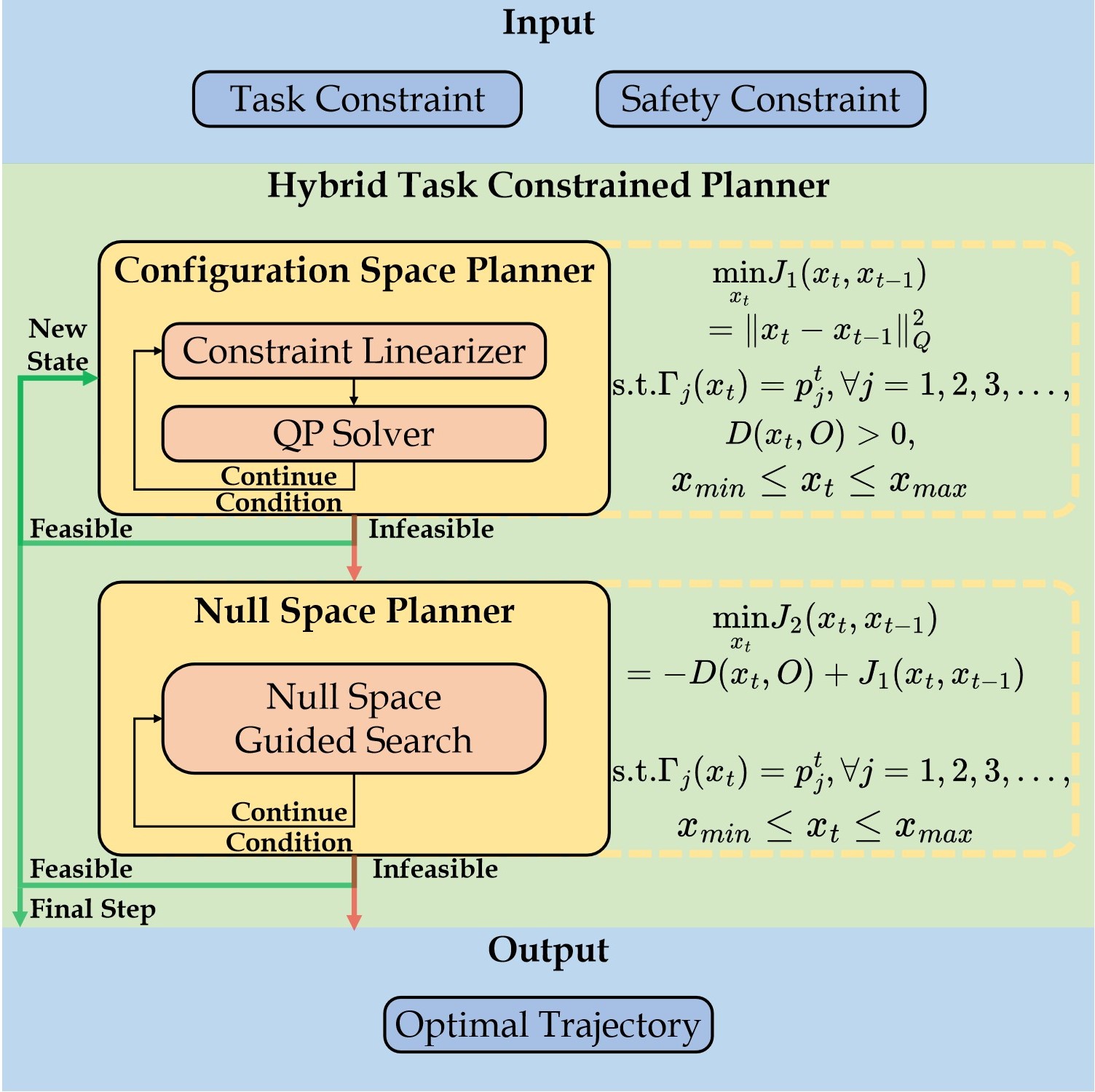}
  \caption{Framework of the hybrid task constrained planner.}
  \label{HTCPFramework}
\end{figure}

\subsection{Iterative Configuration Space Optimization}
\subsubsection{Iterative Equality Linearization Approximation}

To speed up the computation of the highly nonlinear and expensive-to-resolve equality constraint in \eqref{eq:opt1}, we propose iteratively considering the first-order approximation until convergence. We assume $\Gamma$ is twice continuously differentiable, and use a first-order linear approximation to represent the next desired robot body point $C_t$ as:
\begin{align}
    C_t = C_{t-1} + \nabla \Gamma_j(x_{t-1}) \cdot (x_t - x_{t-1}) + \sigma,
    \label{eq:linearization}
\end{align}
where $C_{t-1}$ is the robot body point at state $x_{t-1}$, and $\sigma \to 0$ as $\|x_t - x_{t-1}\|_2^2 \to 0$. This yields a linearized equality constraint, 
\begin{align}
    Jac(x_{t-1})\cdot x_t = Jac(x_{t-1})\cdot x_{t-1} + C_t - C_{t-1} - \sigma \; ,
    \label{eq: linearization final}
\end{align}
for \eqref{eq:opt1}, where $Jac(x) = \nabla \Gamma_j(x)$ is the Jacobian matrix. Solving the optimization problem \eqref{eq:opt1} with this iterative approximation leads to converging results, inspired by iLQR (\cite{Emanuel2004ICINCO}).

\subsubsection{Convex Feasible Set Algorithm}
We use CFS (\cite{liu2018convex}) to efficiently handle the nonlinear inequality constraint \eqref{eq: safety} for safety in confined environments. CFS solves a sequence of convex optimizations constrained in the convex feasible sets to search the non-convex feasible space for solutions. CFS requires that (i) the cost function $J$ is strictly convex and smooth, which is satisfied by \eqref{eq:opt1}, and that (ii) the nonlinear safety inequality constraints can be written as $x\in \Lambda$ where $\Lambda = \cap_i \Lambda_i$ and $\Lambda_i = {x : \phi_i(x) \geq 0}$, $\phi_i$ is a continuous, piecewise and semi-convex smooth function.
For each constraint $\Lambda_i$, We will find a convex feasible set $\mathcal{F}_i$ and construct the convex feasible set around a reference point as $\mathcal{F}(x^r) = \cap_i\mathcal{F}_i(x^r)$. The complete rules of finding $\mathcal{F}_i$ are summarized in (\cite{liu2018convex}), here we approximate $\mathcal{F}_i$ for each constraint $\Lambda_i$ via:
\begin{equation}\label{eq: cfs case 2}
    \mathcal{F}_i(x^r) = \{ x : \phi_i(x^r) + \nabla\phi_i(x^r)(x - x^r) \geq 0 \},
\end{equation}
 with the understanding that the approximation error will be minimized when approaching the optimal solution. This approach works successfully in practice to find optimal solutions that are strictly feasible, as demonstrated in the results section.

\subsubsection{Iterative Convex Optimization in Configuration Space}

After applying the prescribed techniques, we then solve problem \eqref{eq:opt1} via applying CFS algorithm to tackle the inequality constraint \eqref{eq: ineqcon}, and linearizing the nonlinear equality constraint \eqref{eq: eqcon} using \eqref{eq: linearization final}. This results in a quadratic program (QP), which can be iteratively solved as follows:
\begin{equation}
\begin{aligned}
& \underset{x}{\text{min}}
& & J_1(x,x_{ref}) = \| x - x_{ref}\|_Q^2, \\
& \text{s.t.} & &  \nabla D(x_{ref}, O) x \geq \nabla D(x_{ref}, O) x_{ref} - D(x_{ref},O),\\
& & &  Jac(x_{ref})\cdot x = Jac(x_{ref})\cdot x_{ref} + \Delta C - \sigma,\\
& & &  x_{min} \leq x \leq x_{max},\\
\end{aligned}
\label{eq:opt1_new}
\end{equation}
Here, ${\Delta C = C_{next} - C_{ref}}$ represents the difference between the next desired robot body point position $C_{next}$ and the reference robot body point position $C_{ref}$. Initially, we set $x_{ref} = x_{t-1}$ and subsequently update it with the newly computed solution at each iteration. The final solution, obtained when the algorithm converges, is used as the configuration state for the current step.

\subsection{Iterative Null Space Optimization}
The challenge for solving the auxiliary optimization problem \eqref{eq:opt2} lies on dealing with nonlinear equality constraint \eqref{eq: new eq}. Although linearization could deal with \eqref{eq: new eq}, the observation that, \eqref{eq:opt2} is essentially an inverse kinematics problem, inspires us to directly seeks solution in the null space of the redundant DOFs. 

 Formally, when the task constraints dimension ${M}$ is less than the DOFs of robot ${N}$, there exist a null space of the task Jacobian matrix.  Null space is a set of state velocities that results the robot remain to satisfy task equality constraints. In other words, suppose a velocity in null space is $v^*$, then
 \begin{align}
     \Gamma(x) = \Gamma(x + v^* dt),
 \end{align}
 where $dt \rightarrow 0^+$. 
 In order to generate $v^*$ in the null space, we can first solve an orthonormal basis ${N(x_t)}\in \mathbb R^{M \times (N - M)}$ of the null space. $v^*$ can then be represented with ${N(x_t)}*w$ where ${w} \in \mathbb R^{(N - M) \times 1}$ is a coordinate vector.

Therefore, to solve \eqref{eq:opt2}, we can first solve inverse kinematics of \eqref{eq: new eq} for $x_{ref}$, such that $\Gamma_j(x_{ref}) = p_j^t, j = 1,2,3,\cdots$. We then optimize velocity (${N(x_{ref})}*w$) from null space of $x_{ref}$, such that robot can move away from obstacles and is close to $x_{ref}$. Mathematically,
\begin{subequations}
\begin{align}
& \underset{w}{\text{min}}
& & J_2(x, x_{ref}) = -D(x, O) + J_1(x,x_{ref}) \\
& \text{s.t.} & &  \|w\|=1,\\
& & & x = x_{ref} + \alpha*N(x_{ref})*w, \label{eq: updateX}
\end{align}
\label{eq:opt2_new2}
\vspace{-10pt}
\end{subequations}

where $\alpha \rightarrow 0^+$ is a step size. Therefore, \eqref{eq:opt2} can be efficiently tackled via iteratively solving \eqref{eq:opt2_new2}, and update reference configuration accordingly. In practice, $\alpha$ should be non-trivial for computation efficiency, which inevitably introduces deviation to the task equality constraints. Hence, we can re-anchor $x_{ref}$ via additional inverse kinematics computation.  With the implicit form of the task constraint in \eqref{eq:opt2_new2}, we optimize the unit vector ${w}$ instead of the configuration state ${x}$, which reduces the dimension of the optimization problem from ${N}$ to ${(N - M)}$ and improves the computation efficiency.

\subsection{Hybrid Task Constrained Planner}

\begin{algorithm*}[h]
  \caption{Hybrid Task Constrained Planner}\label{HTCP}
  \algblock[TryCatchFinally]{try}{endtry}
  \algcblock[TryCatchFinally]{TryCatchFinally}{finally}{endtry}
  \algcblockdefx[TryCatchFinally]{TryCatchFinally}{catch}{endtry}
	[1]{\textbf{catch} #1}
	{\textbf{end try}}
  \begin{algorithmic}[1]
    \Procedure{Hybrid Task Constrained Planner}{$\mathbb{C}_{target},O,T,x_{ini},\xi$}
      
      \State $x_{pre} = x_{ini}$
	
	\State \textbf{Iteration:}
        \For{$t=0,1,2,..., T$} \Comment{Outer Loop}
      	\State Set $C_{next}\leftarrow\mathbb{C}_{target}(t)$, $x_{ref} \leftarrow x_{pre}$, and $C_{ref} \leftarrow \Gamma(x_{pre})$
       \try
           \State Iteratively solve \eqref{eq:opt1_new} until \eqref{eq: eqcon} and \eqref{eq: ineqcon} are both satisfied \Comment{Configuration Space Planner Inner Loop}
       \catch
            \State Solve \eqref{eq: new eq} to get $x_{ref}$
            \While{$D(x_{ref}, O) < 0$}               \Comment{Null Space Planner Inner Loop}
    		\State Solve the optimization problem \eqref{eq:opt2_new2}, whose solution is ${\hat{w}}$ 
                \State update $x_{ref}$ with  ${\hat{w}}$ using \eqref{eq: updateX} and update $C_{ref}\leftarrow \Gamma(x_{ref})$
    		\State Re-anchor ${x_{ref}}$ with \eqref{eq: new eq} if $\| C_{next} - C_{ref} \| > \xi$
		
    	\EndWhile 
       \endtry
	\State Record $\bm{x}(t) \leftarrow x_{ref}$ and $x_{pre} \leftarrow x_{ref}$
      \EndFor  
      \State \textbf{return} reference trajectory $\bm{x}$
    \EndProcedure
  \end{algorithmic}
  \label{alg1}
\end{algorithm*}

The Hybrid Task Constrained Planner (HTCP) algorithm combines the benefits of \textit{Iterative Configuration Space Optimization} and  \textit{Iterative Null Space Optimization}. It is summarized in Algorithm \ref{alg1}. The input parameters include $\mathbb{C}_{target} \in \mathbb{R}^{M\times T}$, which represents the pre-defined trajectory of the $j$-th robot body point in Cartesian space, $O$ for environment obstacles, $T$ for planning horizon, $x_{ini}$ for the initial robot joint state, and $\xi$ for the equality threshold.

In each outer loop iteration of HTCP, the next desired Cartesian position $C_{next}$ and the configuration $x_{pre}$ from the previous step are obtained. $x_{pre}$ initializes the temporary reference configuration $x_{ref}$ and the corresponding Cartesian position $C_{ref}$, which are updated in multiple rounds. The Configuration Space Planner inner loop procedure starts with the constraint linearizer simplifying the constraints to form the QP problem. The QP solver then solves \eqref{eq:opt1_new} and updates the reference configuration. If the solution is feasible, the planner continues to solve the next desired configuration state. When the solution is infeasible, the Null Space Planner takes control.

The Null Space Planner first tracks $C_{next}$ via solving inverse kinematics and then solves the optimization problem \eqref{eq:opt2_new2} iteratively to get the optimal direction to update the configuration state. After each update, $x_{ref}$ is re-anchored such that task equality constraint is satisfied. This iterative process enables the Null Space Planner to jump out of the local optimal and return control to the Configuration Space Planner for further planning.

\begin{figure}
  \centering
  \begin{subfigure}[t]{0.2\textwidth}
    \includegraphics[width=1.\textwidth]{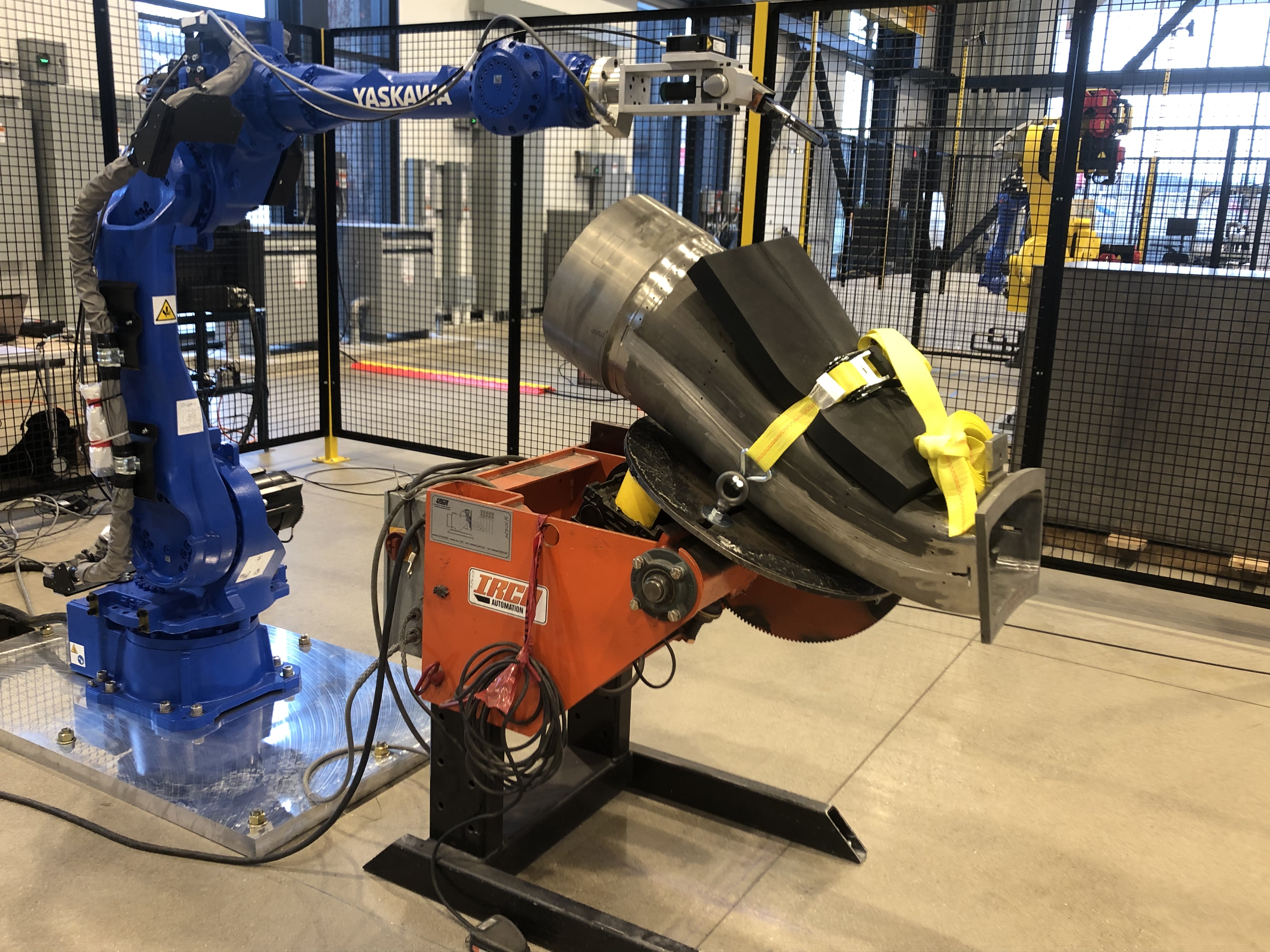}
    \caption{}
    \label{fig: sim a}
  \end{subfigure}
  \begin{subfigure}[t]{0.2\textwidth}
     \includegraphics[width=1.3\textwidth]{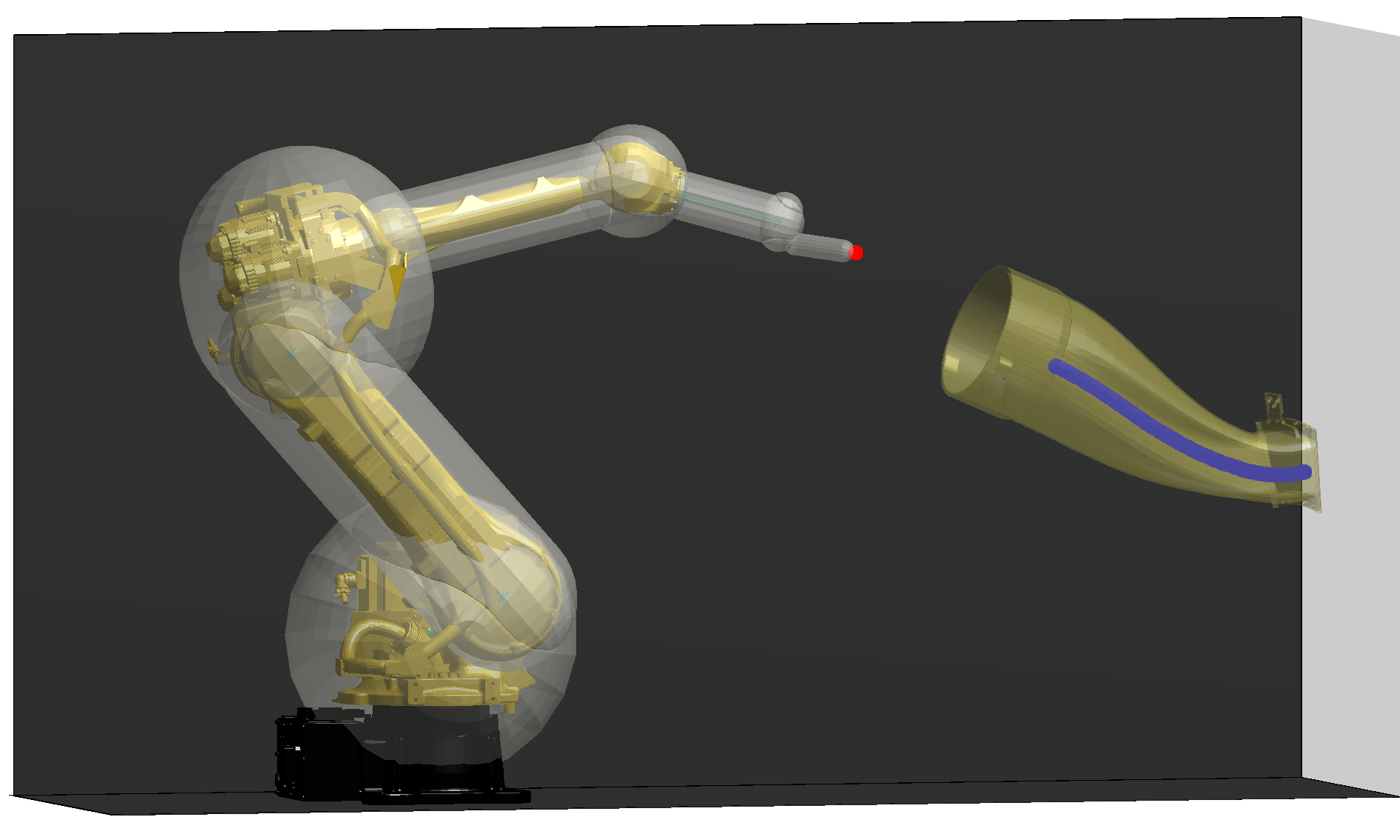}
     \caption{}
     \label{fig: sim b}
  \end{subfigure}
  \caption{Real setup: Blue 6DOF Motoman GP50 robot, orange positioner, silver workpiece with weld inside. Simulator: Gray transparent capsules wrap robot links, blue dots are weld points.}
  \label{setup}
\end{figure}

 \section{Results}
 \subsection{Experimental Setup}
 \begin{wrapfigure}{r}{0.24\textwidth}
\vspace{-10pt}
    \centering
    \includegraphics[width=0.22\textwidth]{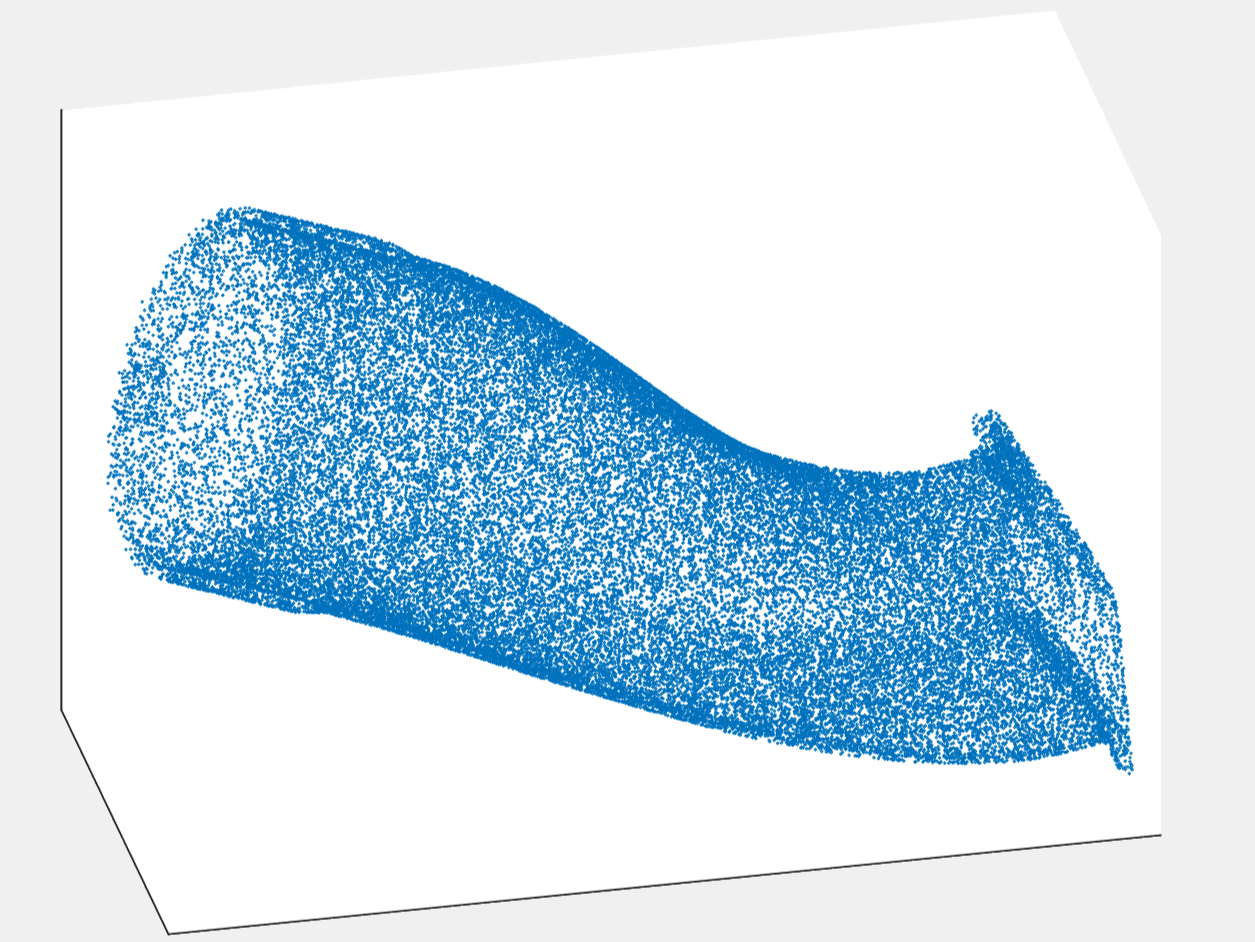}
    \caption{The workpiece obstacle is approximated using a point cloud consists of $50015$ 3-dimension points.}
    \label{pointCloud}
\vspace{-10pt}
\end{wrapfigure}
This section demonstrates the proposed algorithm's effectiveness for a weld grinding task using a 6DOF robot manipulator in confined environments with workspace inequality constraints and task equality constraints. Figure \ref{setup} displays the experimental setup with a YASKAWA Motoman GP50 robot mounted on the ground, and tasked with grinding the weld bead inside the workpiece mounted on the positioner (\cite{zhao2022provably}). 
To simulate the experimental setup, we construct a simulation environment shown in Figure \ref{fig: sim b}, where we simplify the robot links and workpiece's geometry using capsules and a point cloud representation of the workpiece's interior face with 50015 3D points (shown in Figure \ref{pointCloud}), respectively. These representations allow us to define the motion planning constraints as shown below.

\subsection{Task Equality Constraints}
We generate the reference trajectory incrementally using an iterative framework. At time step $t$, the task equality constraint requires that the end-effector tip position computed by forward kinematics with respect to the joint configuration $x_t$ should co-locate with the next desired end-effector tip position (weld point) $C_{next}$, expressed as $\mathbb{FK}(x_t) = C_{next}$. Here, $\mathbb{FK}(x)$ is the forward kinematics function that calculates the end-effector tip position in Cartesian space, and $C_{next} \in \mathbb{R}^3$ represents the desired position of the weld points in Cartesian space.

\subsection{Workspace Inequality Constraints}

At each time step $t$, a task inequality constraint is imposed to ensure the closest distance between the workpiece and robot capsules at configuration $x_t$ to be greater than zero. This is expressed as $D(x_t,O) = \min_j(\mathcal{D}(\Pi_j^t, \mathbb{P})) > 0$, where $\Pi_j^t$ denotes the capsules at time step $t$ with $j=1,2,...,6$, $\mathbb{P}$ is the point cloud, and $\mathcal{D}(\bullet)$ calculates the closest distance between a line segment and the point cloud. The safety constraint does not apply to the end-effector tip, as it needs to maintain contact with the weld bead. Additionally, to ensure that the robot follows the desired path through the workpiece tunnel, a negative distance value is assigned to the segment outside the tunnel, and the inequality constraint requires the distance value to be strictly positive.

\begin{figure}
     \centering
    \begin{subfigure}[t]{0.2\textwidth}
        \raisebox{-\height}{\includegraphics[width=\textwidth]{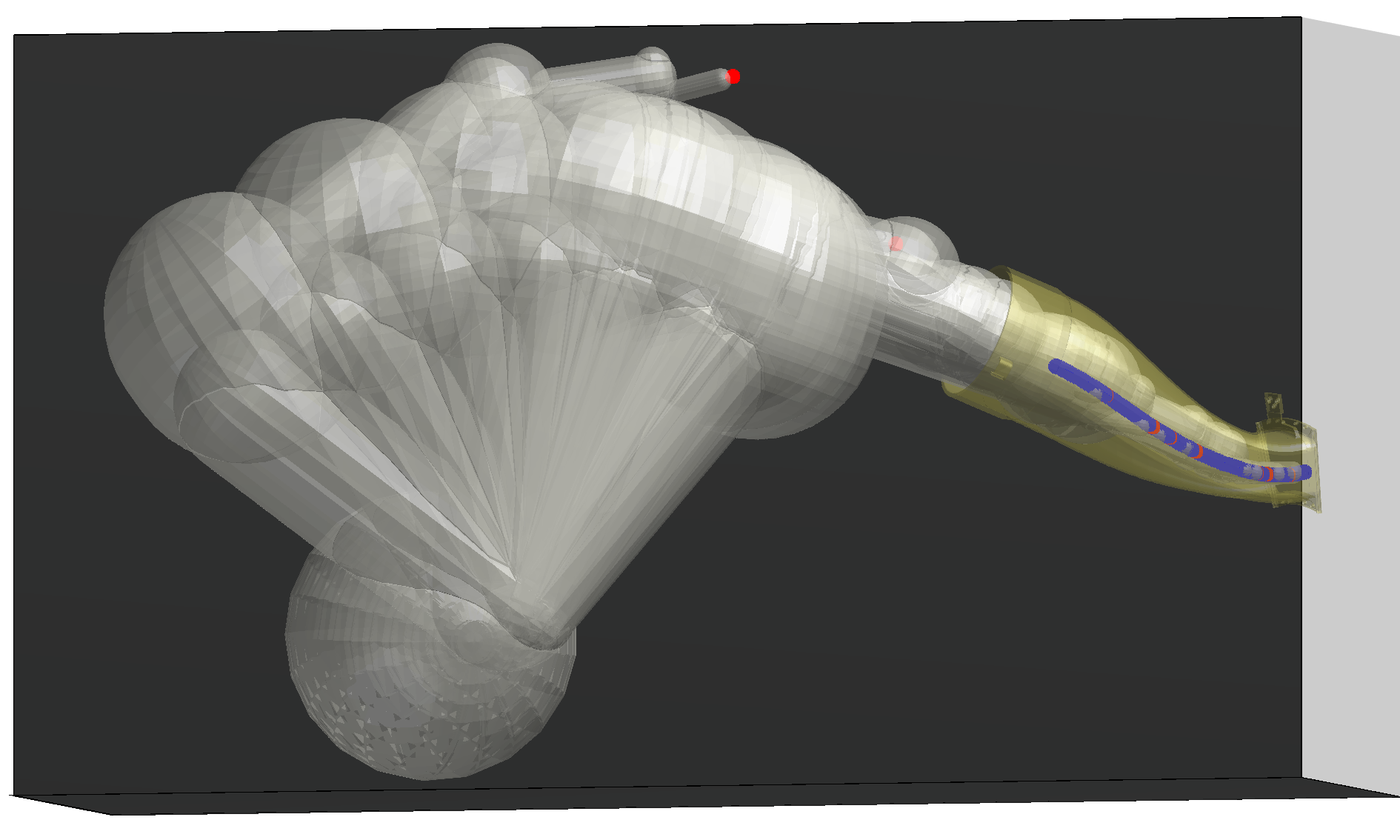}}
    \end{subfigure}
    \begin{subfigure}[t]{0.2\textwidth}
        \raisebox{-\height}{\includegraphics[width=\textwidth]{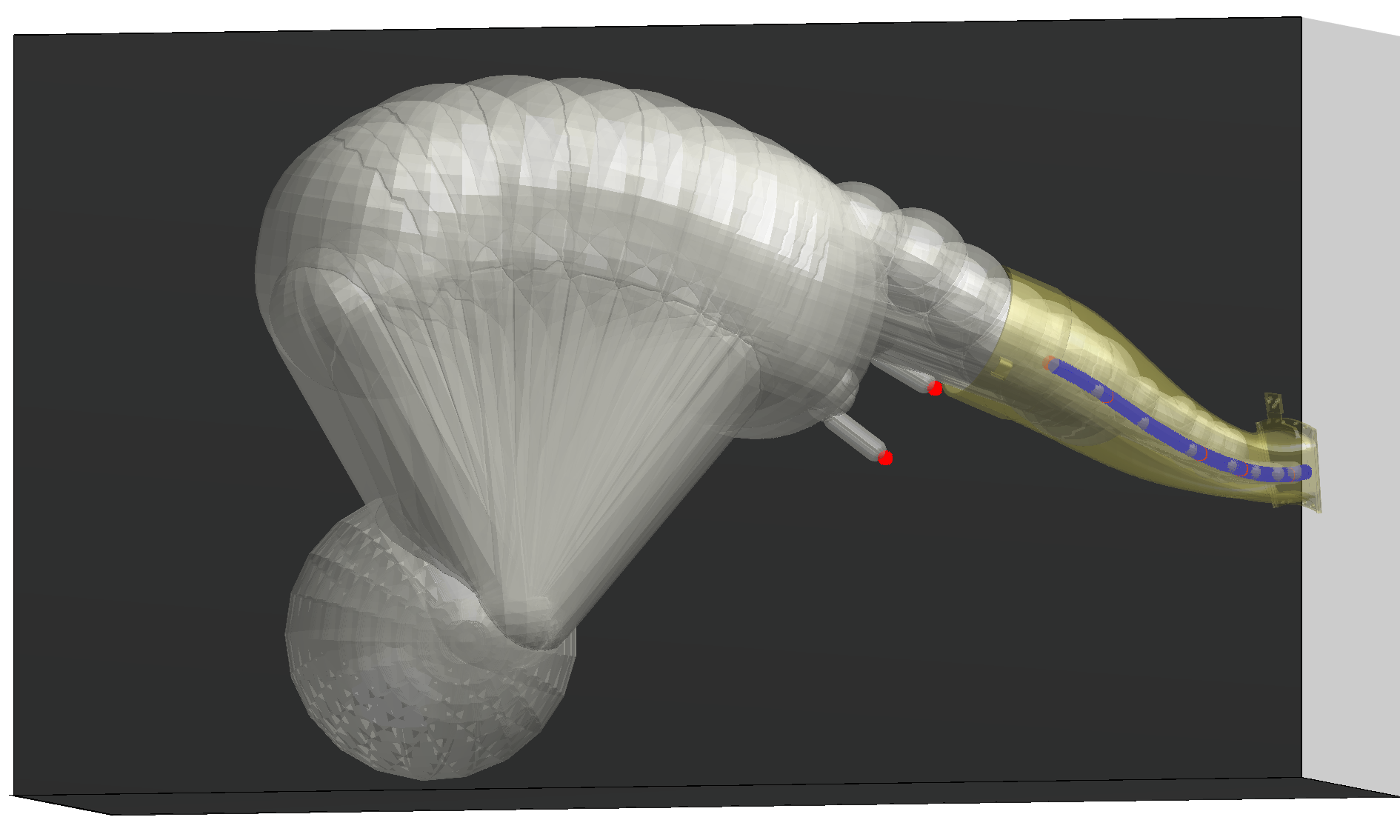}}
    \end{subfigure}
    \begin{subfigure}[t]{0.2\textwidth}
        \raisebox{-\height}{\includegraphics[width=\textwidth]{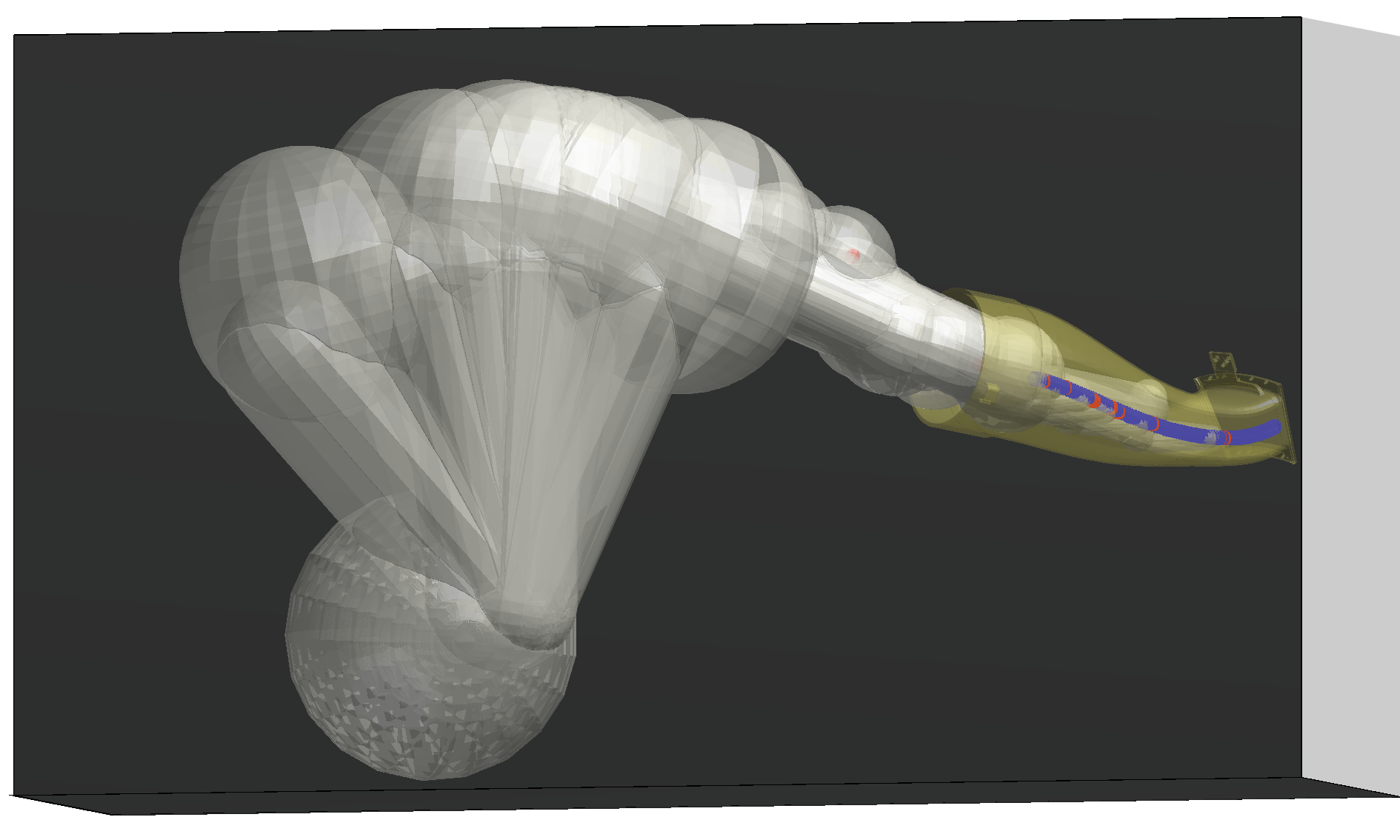}}
    \end{subfigure}
    \begin{subfigure}[t]{0.2\textwidth}
        \raisebox{-\height}{\includegraphics[width=\textwidth]{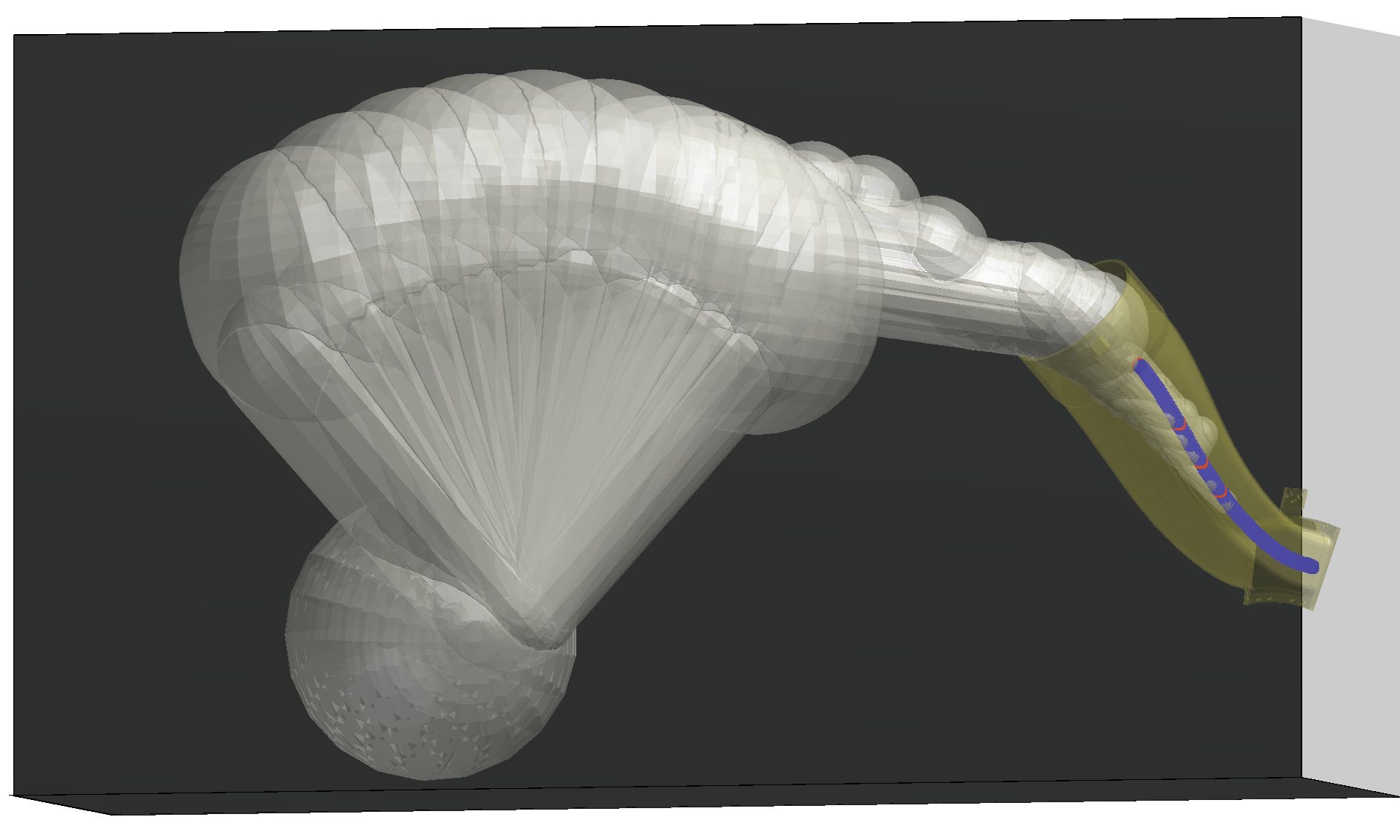}}
    \end{subfigure}
    
     \caption{
    Hybrid task constrained planner generates collision-free polishing trajectories for various robot and workpiece configurations.}
     \label{swipe}
     \vspace{-5pt}
\end{figure}
\begin{table*}
\begin{center}
\captionsetup{width=15cm}
\caption{The polishing trajectory planning performance on 125 robot initial configurations. The average success rate, computation time, TCP distance to weld path, and closest distance between robot and obstacle are reported. The best results are highlighted in bold.}
\label{tb:summarizeRobot}
\begin{tabular}{ccccc}
  & Success rate (\%) & Computation time ($s$) & TCP distance ($mm$) & Safe distance ($mm$)\\\hline
Baseline planner & 66.95 & 0.4281 & 0.4474 & 0.1561\\
C-Space sample & 68.50 & 0.4697 & 0.4510 & 0.1572\\
N-Space sample & 70.85 & 0.5266 & 0.4494 & 0.4622\\
N-Space SQP & 99.84 & \bf{0.4142} & 0.4450 & 0.6309\\
N-Space Interior Point & 99.71 & 0.4145 & \bf{0.4385} & 0.7028\\
N-Space CMA-ES & \bf{100} & 0.6745 & 0.4533 & \bf{0.8200}\\\hline
\end{tabular}
\end{center}
\end{table*}
\subsection{Comparison}

 \begin{table*}
 \centering
 \captionsetup{width=15cm}
\caption{Performance comparison among different methods in terms of generating collision-free polishing trajectory on 49 different workpiece initial configurations. }
\label{tb:summarizeWp}
\begin{tabular}{ccccc}
  & Success rate (\%) & Computation time ($s$) & TCP distance ($mm$) & Safe distance ($mm$)\\\hline
Baseline planner & 39.16 & \bf{0.2789} & 0.4831 & \bf{12.0973}\\
C-Space sample & 49.93 & 0.5740 & 0.4668 & 4.6684\\
N-Space sample & 50.11 & 1.1377 & 0.4557 & 4.1437\\
N-Space SQP & 70.73 & 0.8583 & 0.4207 & 4.3668\\
N-Space Interior Point & \bf{76.87} & 1.0240 & \bf{0.3986} & 1.7241\\
N-Space CMA-ES & 62.22 & 3.2577 & 0.4287 & 7.5280\\\hline
\end{tabular}
\end{table*}

To demonstrate the effectiveness of the proposed method in motion planning for different workpiece and robot configurations, we evaluate the algorithm in a variety of scenarios. The workpiece can be mounted in different positions relative to the robot, and to test this, we perturb the initial workpiece configuration with 49 settings (7 different yaw angles about the $y$ axis of the world frame and 7 different pitch angles about the $z$ axis of the world frame). Additionally, we perturb the robot's initial configuration with 125 different settings (5 different joint positions for joint 3, 4, and 5 respectively).


\subsubsection{Methods for Comparison}
To verify our algorithm's ability to solve the problem of local optima, we compare it against the iterative convex optimization planner as the baseline planner for its efficiency and accuracy. We also develop the following planner for comprehensive evaluation:
\begin{itemize}
    \item \textit{C-Space Sample}: Samples up to 50 different configuration states as new initialization to track the next target point when the planner gets stuck at a local optima.
    \item \textit{N-Space Sample}: Samples up to 50 different null space directions and moves the robot along each of them for 100 steps. The final states are taken as the new initialization to track the next target point when the planner gets stuck at a local optima.
    \item \textit{N-Space SQP}: Applies the null space optimization to solve the configuration state when the planner gets stuck at a local optima. Uses the SQP method to solve the nonlinear optimization problem.
    \item \textit{N-Space Interior Point}: Uses the Interior Point method to solve the same optimization problem as \textit{N-Space SQP}.
    \item \textit{N-Space CMA-ES}: Uses the CMA-ES method to solve the same optimization problem as N-Space SQP, with a population size of 100 and a maximum iteration of 5.
\end{itemize}

\subsubsection{Results}

We compared several task-constrained motion planning algorithms under the same settings for target, obstacles, termination conditions, and optimization objectives. The interior point and SQP algorithms both solved the nonlinear constraints using MATLAB's \verb+fmincon+ function, and all experiments were conducted on MATLAB 2022 with a 2.5 GHz Intel Core i9 processor.

We evaluated the algorithms based on their success rate in tracking the target, average computation time of each step, end-effector distance from the desired weld point (TCP distance), and closest distance between the robot and obstacle (safe distance). We tested a total of $105$ target points, with an equality threshold of $0.001$. The comparison results are summarized in Table \ref{tb:summarizeRobot} and Table \ref{tb:summarizeWp}.

Compared to the baseline planner, both our proposed method and the sampled methods improved the success rate of the solution. However, the sample-based method was less efficient due to the low success rate of random sampling. The C-Space sample had a lower success rate than the N-Space sample, illustrating the importance of reducing the problem dimension for better solutions.

The null space optimization improved the success rate from $66.95\%$ to nearly $100\%$ for different robot configurations, and from $39.16\%$ to around $76.87\%$ for different workpiece configurations. In all experiments, the TCP distance was less than the threshold and the safe distance was greater than zero, indicating that the proposed method satisfied both task and safety constraints.

SQP and Interior Point methods were more stable and efficient compared to the CMA-ES method for null space optimization. The latter highly depends on initialization and can be time-consuming. Additionally, the average computation time cost of SQP and Interior Point-based optimization was comparable to that of the baseline planner, meaning that our proposed method increased success rate without sacrificing efficiency.

\section{Conclusion}


In conclusion, our paper proposed a novel motion planning approach for redundant robot arms using a hybrid optimization framework to search for optimal trajectories in both the configuration space and null space. Our approach efficiently generates high-quality trajectories that satisfy task and collision avoidance constraints while avoiding local optima during incremental planning. Our experiments in an onsite polishing scenario with various robot and workpiece configurations demonstrate significant improvements in trajectory quality compared to existing methods. Our approach has broad applications in industrial tasks involving redundant robot arms. Future work includes incorporating sampling into null space optimization for fully exploring feasible configurations and maximizing trajectory generation potential in confined environments.

\bibliography{ifacconf}             
                                                   
\end{document}